\pdfoutput=1

\documentclass[11pt]{article}

\usepackage[preprint]{acl}
\usepackage{pdfpages} 
\usepackage{pgffor} 
\usepackage{times,latexsym}
\usepackage{url}

\usepackage{hyperref}
\usepackage{enumitem} 
\usepackage{svg}
\usepackage{graphicx}    
\usepackage{float}       
\usepackage{xcolor}
\usepackage{colortbl}
\usepackage{booktabs}
\usepackage{tikz}
\usepackage{longtable}
\usepackage{tcolorbox} 
\usepackage{listings} 
\usepackage{amsmath}
\usepackage{adjustbox}
\usepackage{multirow}
\usepackage{longtable}
\usepackage{lscape}
\definecolor{matplotlibGreen}{HTML}{008000}

\usepackage[T1]{fontenc}

\usepackage[utf8]{inputenc}

\usepackage{microtype}

\usepackage{inconsolata}

\usepackage{graphicx}

\title{Mathematics Isn’t Culture-Free: Probing Cultural Gaps via Entity and Scenario Perturbations}

\author{
 \textbf{Aditya Tomar\textsuperscript{1}},
 \textbf{Nihar Ranjan Sahoo\textsuperscript{1}},
 \textbf{Ashish Mittal\textsuperscript{2}},
 \textbf{Rudra Murthy\textsuperscript{2}},
\\
 \textbf{Pushpak Bhattacharyya\textsuperscript{1}},
\\
 \textsuperscript{1}IIT Bombay,
 \textsuperscript{2}IBM Research, India
\\
\small{\{adityatomar, nihar, pb\}@cse.iitb.ac.in, \{arakeshk, rmurthyv\}@in.ibm.com}
}

\begin{document}
\maketitle
\begin{abstract}
Although mathematics is often considered culturally neutral, the way mathematical problems are presented can carry implicit cultural context. Existing benchmarks like GSM8K are predominantly rooted in Western norms, including names, currencies, and everyday scenarios. In this work, we create culturally adapted variants of the GSM8K test set for five regions \textit{Africa, India, China, Korea, and Japan} using prompt-based transformations followed by manual verification. We evaluate six large language models (LLMs), ranging from 8B to 72B parameters, across five prompting strategies to assess their robustness to cultural variation in math problem presentation. Our findings reveal a consistent performance gap: \textit{models perform best on the original US-centric} dataset and comparatively worse on culturally adapted versions. However, \textit{models with reasoning capabilities are more resilient} to these shifts, suggesting that deeper reasoning helps bridge cultural presentation gaps in mathematical tasks.
\end{abstract}

\section{Introduction}   \label{sec:intro}
Large Language Models (LLMs) have exhibited remarkable capabilities across a wide spectrum of natural language understanding and generation tasks, from open-domain question answering \cite{kamalloo-etal-2023-evaluating} to code generation \cite{code_llm} and multi-step reasoning \cite{multi-step}. Recent advancements have shown that LLMs can achieve near-human performance in solving complex tasks that require logical inference and chain-of-thought reasoning \cite{imitation}.

One task that has garnered particular attention is mathematical problem solving, which serves as a strong proxy for models’ reasoning and symbolic manipulation abilities. Among the benchmarks in this space, GSM8k \cite{gsm8k} has become the de facto standard for evaluating arithmetic and word problem-solving skills in LLMs. Comprising grade-school-level math problems presented in natural language, GSM8k has been used to benchmark a range of models and reasoning techniques, including chain-of-thought prompting \cite{cot}. However, while GSM8k is syntactically diverse, it is culturally homogeneous—nearly all problems are rooted in US-centric scenarios, using American names, dollar-based currency, Western contexts like baseball tickets, etc. So, the \textit{research question} we ask is:
\noindent    \emph{Are LLMs truly reasoning over math, or are they overfitting to culturally familiar problem formats?}\\
To explore this, we propose a systematic cultural adaptation of the GSM8k benchmark. We construct five culturally modified versions of the GSM8k test set corresponding to \textit{India, China, Korea, Japan,} and a \textit{pan-African} context, using a prompt-based rewriting pipeline followed by manual verification. These variants preserve the core mathematical content while altering names, scenarios, and currencies to reflect local cultural norms. We then evaluate six LLMs across five prompting strategies on these datasets.

Our contributions are,
\begin{enumerate}[noitemsep,nolistsep]
\item \textbf{Cultural Benchmark Construction}: We construct a culturally adapted version from the GSM8k test set using a prompt-and-verify pipeline for five different cultures, apart from the original US culture: \textit{African, Indian, Chinese, Korean, Japanese}. (§\ref{sec:dataset}) \footnote{\textit{We will release the dataset and code.}}
\item Comparative performance of six different LLMs across five different prompting setups to assess robustness to cultural shifts in mathematical tasks. We find consistent performance degradation on non-US cultural variants. (§\ref{sec:results})
\end{enumerate}

\section{Related Work}  \label{sec:related}
\begin{figure*}
    \centering
    \includesvg[width=0.9\linewidth]{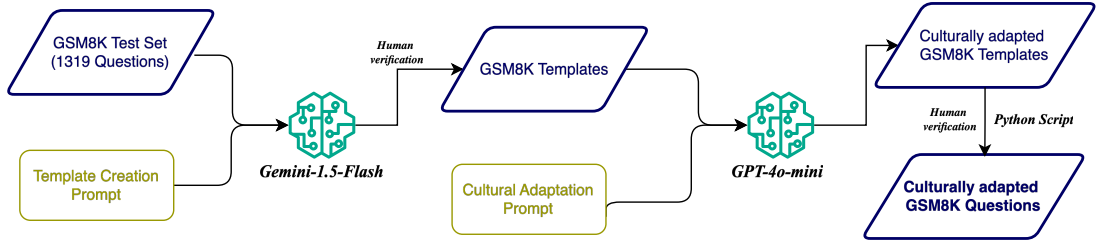}
    \caption{Dataset creation pipeline from US culture to different cultures.}
    \label{fig:dataset-creation-pipeline}
\end{figure*}

Recent advancements in LLMs have significantly improved mathematical reasoning, particularly through CoT prompting \cite{wei2022chain} and benchmarks such as GSM8K. However, models still struggle with symbolic variations and culturally adapted problems \cite{jin2024gsm}. Broader research on cultural bias in NLP reveals that models often reflect Western norms, resulting in reduced performance in unfamiliar cultural settings \cite{blodgett2020language, shah2020predictive, lauscher2020cultural, zhang2022cultural}. In the context of math, \citet{patel2021culturally} demonstrated that culturally framed problems impact model accuracy, revealing a reliance on superficial cues.

\citet{karim2025lostculturaltranslationllms} similarly examined LLMs’ cultural robustness in math by creating six country-specific GSM8K variants using GPT-4o-based symbolic templates and rule-based substitutions (e.g., for Pakistan, Somalia, and Haiti). Their work focuses on surface-level cultural shifts and model performance under a fixed prompt. In contrast, we construct five region-level datasets (\textit{Africa, India, China, Korea, Japan}) through LLM-driven adaptation with manual verification and evaluate multiple prompting strategies across open models to assess cultural resilience.

\section{Dataset}\label{sec:dataset}
To assess the robustness of LLMs to cultural variation in math problem presentation, we create culturally adapted variants of the GSM8k test set. GSM8k is a widely used benchmark comprising 1,319 grade school-level math word problems, originally framed in a predominantly US-centric context. Our culturally adapted dataset spans six regions: the United States (original), India, China, Japan, Korea, and a pan-African category covering diverse African cultural contexts, specifically including Kenya, Tanzania, Morocco, and Nigeria. The complete pipeline of the dataset creation process is shown in Figure \ref{fig:dataset-creation-pipeline}.

\subsection{Template Generation}
We begin by transforming each GSM8k problem into a templated form that abstracts away culturally specific entities. Using the \textit{Gemini-1.5-Flash}\footnote{\href{https://ai.google.dev/gemini-api/docs/models\#gemini-1.5-flash}{models/gemini-1.5-flash}} model, we identify and replace named entities with placeholders. These include person names, locations, currencies, food items, and culturally specific activities, while keeping numerical values and logical structure intact.

\subsection{Cultural Adaptation}
Next, we use GPT-4o-mini\footnote{\href{https://platform.openai.com/docs/models/gpt-4o-mini}{gpt-4o-mini-2024-07-18}} to generate culturally appropriate replacements for each placeholder. For each of the five target regions (India, China, Japan, Korea, and pan-Africa), the model selects named entities that are culturally recognizable and contextually suitable. The values remain fixed to maintain comparability across versions. 

A simple Python script then reconstitutes the adapted problems by injecting the culturally relevant names and terms into the templates. 

An example of the transformation from US culture to Indian culture is shown in  Figure \ref{fig:dataset-example}, and the prompt used is shown in Appendix \ref{app:prompts}.

\subsection{Human Verification}
To ensure semantic fidelity and cross-cultural consistency, all culturally adapted questions were manually verified by annotators\footnote{Annotator demographics are detailed in Appendix \ref{app:annotator}}. This process confirmed that the original problem structure and numerical values were preserved, substituted named entities were culturally appropriate, and no unintended semantic changes occurred. Only problems passing this verification were included. Each regional variant contains 1,319 questions, matching the original GSM8k test set, resulting in a total of 7,914 culturally grounded math problems spanning six cultural contexts.

\section{Experimental Setup}  \label{sec:expt}

In this section, we systematically evaluate how culturally adapted versions of the test set of GSM8k affect the performance of LLMs. Our focus is to quantify performance shifts when modifying contextual entities specific to different cultures while preserving the underlying mathematical structure.

\subsection{Problem Statement}

Let \( Q = \{q_1, q_2, \dots, q_n\} \) be the set of original GSM8K math word problems rooted in Western cultural contexts. For each culture \( c \in \mathcal{C} = \{\text{Indian}, \text{Chinese}, \text{Japanese}, \text{Korean}, \text{African}\} \), we define a culturally adapted test set \( Q^c = \{q_1^c, q_2^c, \dots, q_n^c\} \), where each \( q_i^c \) is a semantically equivalent reformulation of \( q_i \), differing only in surface-level cultural cues (e.g., names, currencies, scenarios).

Given an LLM \( M \) and a prompting strategy \( P \), we denote its accuracy on \( Q \) as \( \text{Acc}(M, P, Q) \), and on the culturally adapted version as \( \text{Acc}(M, P, Q^c) \). The core problem is to quantify and statistically evaluate the performance gap:
\[
\Delta_c = \text{Acc}(M, P, Q) - \text{Acc}(M, P, Q^c)
\]
and determine whether \( \Delta_c > 0 \) is statistically significant, suggesting cultural sensitivity in mathematical reasoning performance.

\subsection{Models Evaluated}
We evaluate a set of six open-source instruction-tuned LLMs, spanning a range from 8B to 72B parameters and varying architecture:
LLaMA 3.1–8B–Instruct, LLaMA 3.1–70B–Instruct, Gemma 2–9B–it, Gemma 3–27B–it, Mixtral 8×7B–Instruct v0.1, and Qwen 2.5–72B–Instruct\footnote{\url{meta-llama/Llama-3.1-8B-Instruct}, \url{meta-llama/Llama-3.1-70B-Instruct}, \url{google/gemma-2-9b-it}, \url{google/gemma-3-27b-it}, \url{mistralai/Mixtral-8x7B-Instruct-v0.1}, \url{Qwen/Qwen2.5-72B-Instruct}}.

\subsection{Prompting Strategies} \label{sec:prompt}

We employ five prompting strategies for each model and culture pair, each reflecting different levels of supervision and guidance in solving math problems: \textit{zero-shot}, \textit{zero-shot chain-of-thought} (CoT), \textit{one-shot}, \textit{one-shot CoT}, \textit{chain-of-draft} \cite{xu2025chaindraftthinkingfaster}. The exact prompts and the one-shot example used for each method are presented in the Appendix \ref{app:prompts}.

\subsection{Hypothesis Testing}

To assess whether performance differences between the US GSM8K and its culturally adapted versions are statistically significant, McNemar’s test \cite{McNemar1947} is applied to matched question pairs. Further details are provided in Appendix \ref{app:testing}.




\section{Results}\label{sec:results}
\begin{figure*}[t]
    \centering
    \includesvg[width=0.95\linewidth]{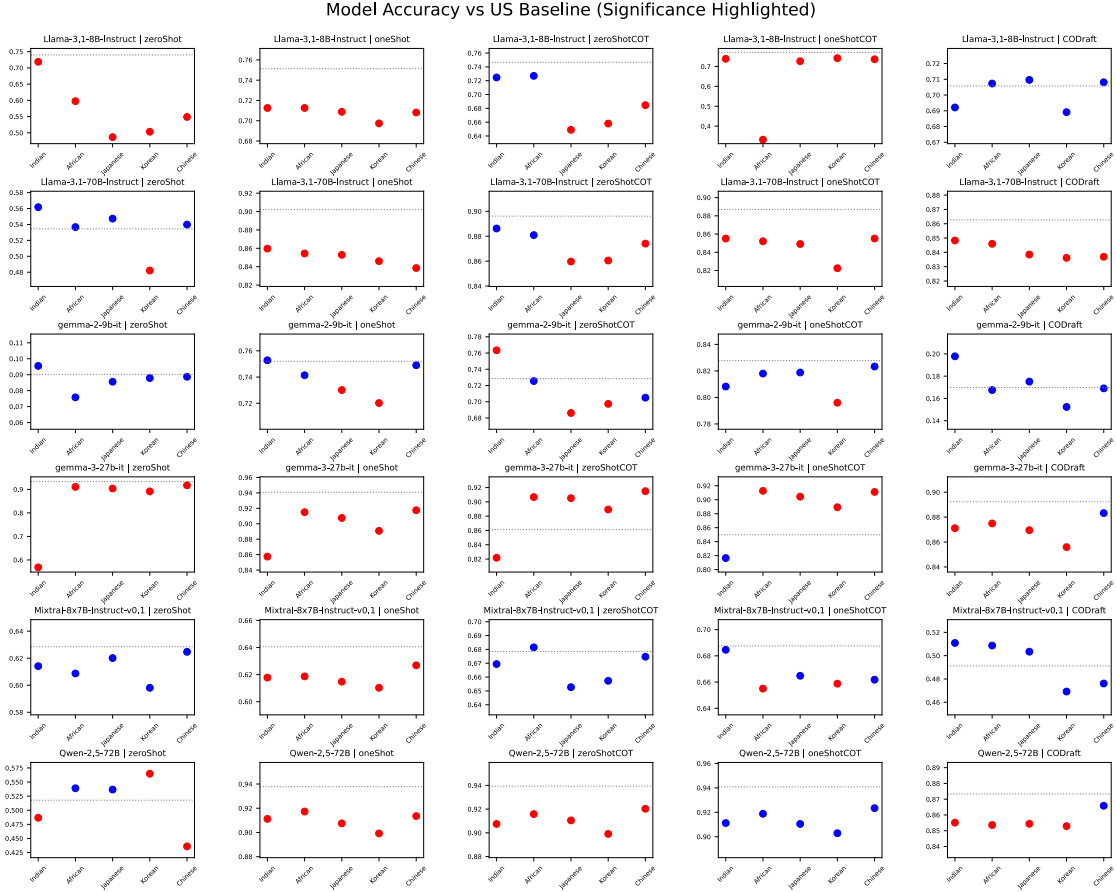}
    \caption{Model accuracy across culturally adapted GSM8k datasets relative to the US baseline. Each subplot shows the accuracy of a specific model and prompting technique across five cultural variants: \textit{Indian, Korean, Chinese, Japanese}, and \textit{African}. The \textit{dashed horizontal line} (....) represents the model’s accuracy on the original US-context GSM8k dataset.
    \textcolor{red}{Red dot} indicates statistically significant differences from the US baseline, while \textcolor{blue}{blue dot} denotes non-significant differences.}
    \label{fig:culture-acc-diff}
\end{figure*}

To evaluate the cultural robustness of LLMs, we measured their performance on culturally adapted versions of the GSM8k dataset, spanning five distinct cultural contexts: Indian, Korean, Chinese, Japanese, and African. The results are presented in Figure \ref{fig:culture-acc-diff} and Figure \ref{fig:mean-acc-diff}, which together provide a comprehensive view of performance variations across models, prompt styles, and cultures.

Figure \ref{fig:culture-acc-diff} displays the accuracy of each model under five prompting techniques: zero-shot, one-shot, one-shot CoT, zero-shot CoT, and Chain-of-Draft, compared to the model’s baseline performance on the US version of the dataset. Red markers indicate a statistically significant difference in accuracy from the US baseline, while blue markers denote non-significant differences. Notably, certain models, such as LLaMA-3.1-8B-Instruct, exhibited consistent performance drops across all cultural variants and prompt types, with many of these differences being statistically significant. LLaMA-3.1-70B-Instruct demonstrated more stable performance in zero-shot prompts, with fewer significant deviations, but the difference was significant for one-shot, one-shot-CoT, and COD.

Prompting strategies also influenced model robustness. CoT-based prompts, especially one-shot CoT, tended to reduce performance gaps in some models, suggesting that explicit reasoning steps may help bridge cultural context shifts. However, this effect was not uniform; for example, the Mixtral-8x7B model showed relatively consistent performance across all prompting methods except one-shot, with only minor cultural degradation, while Gemma-3-27B-it showed more sensitivity across all the prompts.

However, the Gemma-3-27B-it model is notably not prone to cultural variation; in fact, for the African, Chinese, and Japanese variants, it even shows better performance than the US baseline when considering the mean accuracy across all prompting techniques. This suggests a higher degree of cultural robustness in Gemma-3-27B-it, potentially reflecting more diverse training data or architectural advantages.

To summarize these findings, Figure \ref{fig:mean-acc-diff} displays a heatmap showing the average accuracy difference from the US baseline for each model–culture pair, aggregated across all prompting techniques. The largest accuracy drops occurred with LLaMA-3.1-8B-Instruct, especially on the African (–0.13) and Japanese (–0.087) variants. Conversely, Gemma-3-27B-it demonstrated strong resilience to cultural shifts, with some minor accuracy gains in certain cases (e.g., +0.013 for the Chinese variant). This indicates that some models may be naturally more robust to cultural context changes, possibly due to factors like training data diversity or model architecture.

Table \ref{tab:reasoning-example-err} provides an example where Mixtral-8x7B-Instruct-v0.1 answered correctly for the US culture but failed on another cultural variant. For full results, see Table \ref{tab:results-num}.

\section{Summary, Conclusion and Future Work}  \label{sec:conclusion}
This study reveals that LLM performance on math reasoning tasks is sensitive to cultural context, with notable accuracy drops in certain regions, particularly for models like LLaMA-3.1-8B-Instruct. In contrast, models such as Gemma-3-27B-it and Mixtral-8x7B-Instruct-v0.1 showed stronger cross-cultural resilience, influenced by factors like training diversity and architecture. Prompting techniques, especially chain-of-thought, helped reduce some cultural performance gaps.

The results highlight the need for culturally adaptable LLMs and emphasize the importance of fine-grained cultural representation, especially within broad categories like pan-African, during development and evaluation. Ensuring equitable global performance will require culturally grounded benchmarks and adaptation strategies.

\section*{Limitation}
Our study presents several limitations that should be considered when interpreting the results. First, we focus exclusively on open-source models and do not evaluate proprietary or closed-source models such as GPT-4 or Claude, which may exhibit different behaviors in cross-cultural mathematical reasoning. This choice was primarily driven by the lack of access to these systems’ internal mechanisms and the constraints associated with reproducibility. Second, due to resource constraints, particularly GPU availability, we limit our evaluation to six models that represent a diverse but selective subset of the open-source LLM landscape. While these models were carefully chosen to cover a range of architectures and sizes, a broader evaluation might yield additional insights. Lastly, all experiments were conducted on NVIDIA A100 GPUs, which, while powerful, imposed practical limitations on the scale and frequency of evaluations, especially for larger models and more complex prompting strategies. Future work can extend this study by incorporating closed-source models, expanding the number of evaluated models, and exploring cultural robustness across additional computational settings.


\bibliography{custom}

\begin{thebibliography}{16}
\providecommand{\natexlab}[1]{#1}

\bibitem[{Blodgett et~al.(2020)Blodgett, Barocas, Daumé~III, and Wallach}]{blodgett2020language}
Su~Lin Blodgett, Solon Barocas, Hal Daumé~III, and Hanna Wallach. 2020.
\newblock Language (technology) is power: A critical survey of "bias" in nlp.
\newblock \emph{arXiv preprint arXiv:2005.14050}.

\bibitem[{Chen et~al.(2021)Chen, Tworek, and co.}]{code_llm}
Mark Chen, Jerry Tworek, and co. 2021.
\newblock \href {https://arxiv.org/abs/2107.03374} {Evaluating large language models trained on code}.
\newblock \emph{Preprint}, arXiv:2107.03374.

\bibitem[{Cobbe et~al.(2021)Cobbe, Kosaraju, Bavarian, Chen, Jun, Kaiser, Plappert, Tworek, Hilton, Nakano, Hesse, and Schulman}]{gsm8k}
Karl Cobbe, Vineet Kosaraju, Mohammad Bavarian, Mark Chen, Heewoo Jun, Lukasz Kaiser, Matthias Plappert, Jerry Tworek, Jacob Hilton, Reiichiro Nakano, Christopher Hesse, and John Schulman. 2021.
\newblock \href {https://arxiv.org/abs/2110.14168} {Training verifiers to solve math word problems}.
\newblock \emph{Preprint}, arXiv:2110.14168.

\bibitem[{Jin et~al.(2024)Jin, Wu, Li, Yang, Sikarwar, Mai, Chang, Qin, Hovy, He et~al.}]{jin2024gsm}
Jie Jin, Zhangyue Wu, Yujie Li, Zixing Yang, Abhineet Sikarwar, Bowen Mai, Quanyu Chang, Qiuyuan Qin, Eduard Hovy, Pengcheng He, and 1 others. 2024.
\newblock Gsm-symbolic: Understanding the limitations of mathematical reasoning in large language models.
\newblock \emph{arXiv preprint arXiv:2410.05229}.

\bibitem[{Kamalloo et~al.(2023)Kamalloo, Dziri, Clarke, and Rafiei}]{kamalloo-etal-2023-evaluating}
Ehsan Kamalloo, Nouha Dziri, Charles Clarke, and Davood Rafiei. 2023.
\newblock \href {https://doi.org/10.18653/v1/2023.acl-long.307} {Evaluating open-domain question answering in the era of large language models}.
\newblock In \emph{Proceedings of the 61st Annual Meeting of the Association for Computational Linguistics (Volume 1: Long Papers)}, pages 5591--5606, Toronto, Canada. Association for Computational Linguistics.

\bibitem[{Karim et~al.(2025)Karim, Karim, Lohana, Keon, Singh, and Sattar}]{karim2025lostculturaltranslationllms}
Aabid Karim, Abdul Karim, Bhoomika Lohana, Matt Keon, Jaswinder Singh, and Abdul Sattar. 2025.
\newblock \href {https://arxiv.org/abs/2503.18018} {Lost in cultural translation: Do llms struggle with math across cultural contexts?}
\newblock \emph{Preprint}, arXiv:2503.18018.

\bibitem[{Lauscher et~al.(2020)Lauscher, Ravishankar, Vulić, and Glavaš}]{lauscher2020cultural}
Anne Lauscher, Vinit Ravishankar, Ivan Vulić, and Goran Glavaš. 2020.
\newblock From zero to hero: On the limitations of zero-shot language transfer with multilingual transformers.
\newblock In \emph{Proceedings of the 2020 Conference on Empirical Methods in Natural Language Processing (EMNLP)}, pages 4483--4499.

\bibitem[{McNemar(1947)}]{McNemar1947}
Quinn McNemar. 1947.
\newblock \href {https://doi.org/10.1007/bf02295996} {Note on the sampling error of the difference between correlated proportions or percentages}.
\newblock \emph{Psychometrika}, 12(2):153--157.

\bibitem[{Patel and Pavlick(2021)}]{patel2021culturally}
Krishna Patel and Ellie Pavlick. 2021.
\newblock Culturally relevant math problems: Investigating machine learning's response to diversity in education.
\newblock \emph{arXiv preprint arXiv:2107.04964}.

\bibitem[{Shah et~al.(2020)Shah, Schwartz, and Hovy}]{shah2020predictive}
Deven Shah, H~Andrew Schwartz, and Dirk Hovy. 2020.
\newblock Predictive biases in natural language processing models: A conceptual framework and overview.
\newblock In \emph{Proceedings of the 58th Annual Meeting of the Association for Computational Linguistics}, pages 5248--5264.

\bibitem[{Srivastava et~al.(2023)Srivastava, Rastogi, Rao, Shoeb, and co.}]{imitation}
Aarohi Srivastava, Abhinav Rastogi, Abhishek Rao, Abu Awal~Md Shoeb, and co. 2023.
\newblock \href {https://arxiv.org/abs/2206.04615} {Beyond the imitation game: Quantifying and extrapolating the capabilities of language models}.
\newblock \emph{Preprint}, arXiv:2206.04615.

\bibitem[{Wei et~al.(2022)Wei, Wang, Schuurmans, Bosma, Ichter, Xia, Chi, Le, and Zhou}]{wei2022chain}
Jason Wei, Xuezhi Wang, Dale Schuurmans, Maarten Bosma, Brian Ichter, Fei Xia, Ed~Chi, Quoc Le, and Denny Zhou. 2022.
\newblock Chain-of-thought prompting elicits reasoning in large language models.
\newblock \emph{arXiv preprint arXiv:2201.11903}.

\bibitem[{Wei et~al.(2023)Wei, Wang, Schuurmans, Bosma, Ichter, Xia, Chi, Le, and Zhou}]{cot}
Jason Wei, Xuezhi Wang, Dale Schuurmans, Maarten Bosma, Brian Ichter, Fei Xia, Ed~Chi, Quoc Le, and Denny Zhou. 2023.
\newblock \href {https://arxiv.org/abs/2201.11903} {Chain-of-thought prompting elicits reasoning in large language models}.
\newblock \emph{Preprint}, arXiv:2201.11903.

\bibitem[{Wei et~al.(2025)Wei, Liu, Wu, and Fang}]{multi-step}
Ting-Ruen Wei, Haowei Liu, Xuyang Wu, and Yi~Fang. 2025.
\newblock \href {https://arxiv.org/abs/2502.14333} {A survey on feedback-based multi-step reasoning for large language models on mathematics}.
\newblock \emph{Preprint}, arXiv:2502.14333.

\bibitem[{Xu et~al.(2025)Xu, Xie, Zhao, and He}]{xu2025chaindraftthinkingfaster}
Silei Xu, Wenhao Xie, Lingxiao Zhao, and Pengcheng He. 2025.
\newblock \href {https://arxiv.org/abs/2502.18600} {Chain of draft: Thinking faster by writing less}.
\newblock \emph{Preprint}, arXiv:2502.18600.

\bibitem[{Zhang et~al.(2022)Zhang, Smith, Baldwin, and Plank}]{zhang2022cultural}
Alice Zhang, Noah~A Smith, Tim Baldwin, and Barbara Plank. 2022.
\newblock Cultural adaptation of language models.
\newblock \emph{arXiv preprint arXiv:2209.11390}.

\end{thebibliography}

\appendix

\begin{figure*}
    \centering
    \includesvg[width=\linewidth]{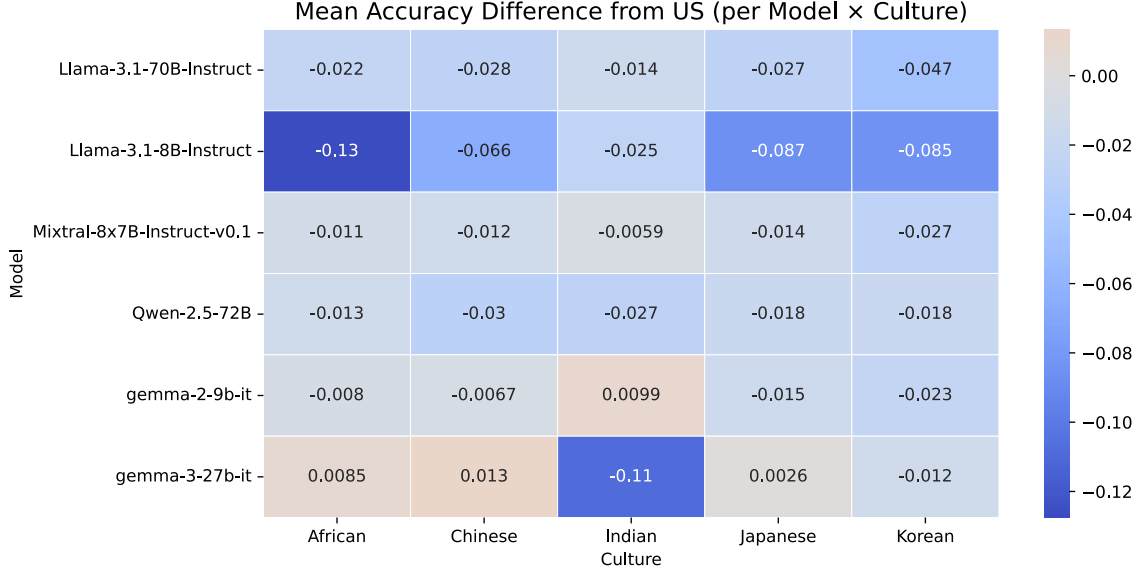}
    \caption{Mean accuracy difference from the US baseline across models and cultural variants.}
    \label{fig:mean-acc-diff}
\end{figure*}

\begin{figure*}
    \centering
    \includesvg[width=\linewidth]{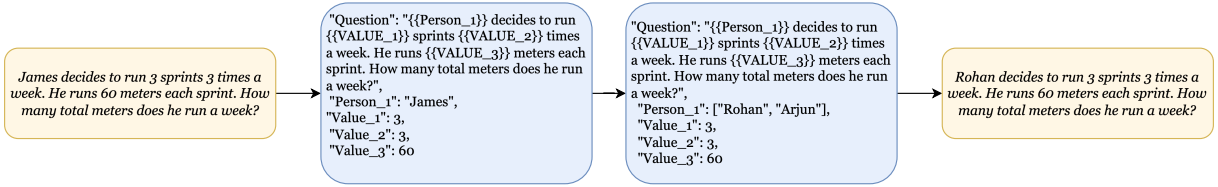}
    \caption{Example showing transformation of the original GSM8k question in US culture to Indian culture.}
    \label{fig:dataset-example}
\end{figure*}

\begin{table*}[ht]
    \centering
    \begin{tabular}{|p{0.08\linewidth}|p{0.8\linewidth}|p{0.08\linewidth}|}
    \hline
    \textbf{Question} & \textbf{Jim spends 2 hours watching TV and then decides to go to bed and reads for half as long. He does this 3 times a week. How many hours does he spend on TV and reading in 4 weeks?} & \textbf{Answer} \\
    \hline
    \textbf{US} & Jim spends 2 hours watching TV and half of that time reading each time he does this routine. So, in one occurrence, he spends \(2 + \frac{1}{2} \times 2 = 3\) hours. He does this 3 times a week, so in one week, he spends \(3 \times 3 = 9\) hours. In 4 weeks, he spends \(9 \times 4 = 36\) hours. \#\#\#\#36 & 36 \\
    \hline
    \textbf{Indian} & In one week, Rohan spends 2 hours (TV) + \(2/2 = 1\) hour (reading) = 3 hours on TV and reading. In four weeks, he spends \(3 \times 4 = 12\) hours. On TV, he spends \(2 \times 4 = 8\) hours. On reading, he spends \(1 \times 4 = 4\) hours. \#\#\#\#Rohan spends 8 hours on TV and 4 hours on reading in 4 weeks. & 8 \\
    \hline
    \textbf{African} & In one week, Musa spends 2 hours (TV) + 2 hours/2 (reading) = 2 + 1 = 3 hours on TV and reading. In four weeks, Musa spends 3 hours/week \* 4 weeks = 12 hours on TV and reading. To find out how much time is spent on TV and reading separately, Musa spends 2 hours/week on TV and 1 hour/week on reading. So, in four weeks, Musa spends 2 hours/week \* 4 weeks = 8 hours watching TV, and 1 hour/week \* 4 weeks = 4 hours reading. \#\#\#\#8 hours for TV and 4 hours for reading. & 8 \\
    \hline
    \textbf{Chinese} & In one week, Li spends 2 hours (TV time) + 2 hours/2 (reading time) = 2 + 1 = 3 hours. So in 4 weeks, Li spends 3 hours/week \* 4 weeks = 12 hours. For TV time only, Li spends 2 hours/week \* 4 weeks = 8 hours. For reading time only, Li spends 1 hour/week \* 4 weeks = 4 hours. \#\#\#\# Li spends a total of 12 hours (8 hours on TV and 4 hours on reading) in 4 weeks. & 12 \\
    \hline
    \textbf{Korean} & In one week, Sujin spends 2 hours watching TV and 2/2 * 1/2 = 1 hour reading. So in one week, Sujin spends 2 + 1 = 3 hours in total. In 4 weeks, Sujin spends 3 * 4 = 12 hours in total. \#\#\#\#12 & 12 \\
    \hline
    \textbf{Japanese} & In one week, Ken spends 2 hours watching TV and 2/2 * 1/2 = 1 hour reading. In one week, Ken spends 2 + 1 = 3 hours in total. In four weeks, Ken spends 3 * 4 = 12 hours in total. \#\#\#\#12 & 12 \\
    \hline
    \end{tabular}
    \caption{An Example of GSM8K-Style Arithmetic Reasoning Across Cultures in a One-Shot Setting Using Mixtral-8x7B-Instruct-v0.1.}
    \label{tab:reasoning-example-err}
\end{table*}

\section{Annotator Demographics}  \label{app:annotator}
We employ three annotators to verify the cultural translations to check any inconsistencies, numerical errors, etc. All the annotators were male computer science graduate students of Indian origin.

\section{Prompts Used}
\label{app:prompts}

We present the exact prompt templates along with the one-shot example (whenever applicable) used for each prompting strategy described in Section~\ref{sec:prompt}.

\begin{tcolorbox}[colback=gray!5!white, colframe=black, title=Cultural Adaptation Prompt]
Given the question template in JSON format with keys and values. First check if the named entities are applicable in \{CULTURE\_NAME\} context if they are not applicable in \{CULTURE\_NAME\} context then change it into the given context.\\
Only change the named entities such as name of a person, currency, units, etc. Do not change the values.\\
Don't generate any comment.\\
\texttt{<CULTURE\_SPECIFIC\_EXAMPLE\_TEMPLATE>}
\end{tcolorbox}

\begin{tcolorbox}[colback=gray!5!white,colframe=gray!75!black,title=ZeroShot Prompt]
Solve the following math word problem and give the answer after separator \#\#\#\#.\\
Don't generate extra examples.\\
Question: \{question\}\\
Answer:
\end{tcolorbox}

\begin{tcolorbox}[colback=gray!5!white,colframe=gray!75!black,title=OneShot Prompt]
Solve the following math word problem and give the answer after separator \#\#\#\#.\\
Don't generate extra examples.\\
Question: There are 3 balls in box 1 and 2 balls in box 2. How many balls are there?\\
Answer: Box 1 contains 3 balls, box 2 contains 2 balls.\\
Total balls = 3 + 2 = 5.\\
\#\#\#\#5\\
Question: \{question\}\\
Answer:
\end{tcolorbox}

\begin{tcolorbox}[colback=gray!5!white,colframe=gray!75!black,title=ZeroShot Chain-of-Thought Prompt]
Think step by step and generate the numerical answer after the separator \#\#\#\#.\\
Don't generate extra examples.\\
Question: \{question\}\\
Answer:
\end{tcolorbox}

\begin{tcolorbox}[colback=gray!5!white,colframe=gray!75!black,title=OneShot Chain-of-Thought Prompt]
Think step by step and generate the numerical answer after the separator \#\#\#\#.\\
Don't generate extra examples.\\
Question: There are 3 balls in box 1 and 2 balls in box 2. How many balls are there?\\
Answer: Step1: Box 1 contains 3 balls.\\
Step2: Box 2 contains 2 balls.\\
Step3: Total balls in Box 1 and Box 2 = 3 + 2 = 5.\\
\#\#\#\#5\\
Question: \{question\}\\
Answer:
\end{tcolorbox}

\begin{tcolorbox}[colback=gray!5!white,colframe=gray!75!black,title=Chain-of-Draft Prompt]
Think step by step, but only keep a minimum draft for each thinking step, with 5 words at most.\\
Return the answer at the end of the response after a separator \#\#\#\#.\\
Question: There are 3 balls in box 1 and 2 balls in box 2. How many balls are there?\\
Answer: x = 3; y = 2; x + y = 3 + 2 = 5. \#\#\#\# 5\\
Question: \{question\}\\
Answer:
\end{tcolorbox}

\section{McNemar's Test for Hypothesis Testing}  \label{app:testing}

To assess whether the performance of language models differs significantly between the original GSM8k (US) and its culturally adapted versions, we conduct hypothesis testing using McNemar’s Test, a non-parametric method for paired nominal data.

\subsection*{Test Motivation and Setup}

For each culture \( c \in \{\text{Indian}, \text{Chinese}, \text{Japanese}, \text{Korean}, \text{African}\} \), we compare model predictions on a shared set of problem instances from:
\begin{itemize}
    \item The original GSM8k dataset (\( Q^{\text{US}} \))
    \item The culturally adapted dataset (\( Q^c \))
\end{itemize}

Since each question \( q_i \) appears in both versions with only surface-level cultural modifications (e.g., names, currency, context), we treat them as paired samples and compare correctness labels of the model’s predictions.

\subsection*{McNemar’s Test}
McNemar’s Test evaluates whether the marginal frequencies of two related binary outcomes are significantly different. Let:
\begin{itemize}
    \item \( n_{01} \): Number of questions the model got wrong in \( Q^{\text{US}} \) but right in \( Q^c \)
    \item \( n_{10} \): Number of questions the model got right in \( Q^{\text{US}} \) but wrong in \( Q^c \)
\end{itemize}

The McNemar test statistic is computed as:
\[
\chi^2 = \frac{( |n_{01} - n_{10}| - 1 )^2}{n_{01} + n_{10}}
\]

This statistic follows a chi-squared distribution with 1 degree of freedom. We use a continuity correction by subtracting 1 from the numerator, which is standard when sample sizes are small.

\subsection*{Hypotheses}
\begin{itemize}
    \item \textbf{Null Hypothesis (H\textsubscript{0})}: Cultural adaptation does not affect model accuracy. The probabilities of a correct answer are equal across both conditions.
    \item \textbf{Alternate Hypothesis (H\textsubscript{1})}: The model is more likely to answer correctly on the original GSM8K question than on its culturally adapted version.
\end{itemize}

\subsection*{Evaluation Criteria}
A prediction is considered correct if it matches the ground truth up to a numerical tolerance of \(10^{-3}\), accounting for minor floating-point discrepancies.

We report the values of \( n_{01} \), \( n_{10} \), the McNemar test statistic, and the corresponding \( p \)-value. A significance threshold of \( \alpha = 0.05 \) is used. If the resulting \( p \)-value is less than 0.05, we reject the null hypothesis and conclude that performance differences due to cultural adaptation are statistically significant.

\subsection*{Implementation Notes}
We compute the contingency table on a per-model, per-prompt, per-culture basis. Detailed results are presented in Section~\ref{sec:results}.

\begin{table*}[htbp]

\centering
\begin{adjustbox}{width=\textwidth}
\begin{tabular}{llcccccc}
\toprule
\textbf{Prompt} & \textbf{Model} & \textbf{US} & \textbf{Indian} & \textbf{African} & \textbf{Japanese} & \textbf{Korean} & \textbf{Chinese} \\
\midrule

\multirow{6}{*}{Zero-shot}
 & Llama-3.1-8B-Instruct &  0.7399 & \cellcolor{blue!25}0.7187 & \cellcolor{blue!25}0.5975 & \cellcolor{blue!25}0.4867 & \cellcolor{blue!25}0.5034 & \cellcolor{blue!25}0.5489  \\
 & Llama-3.1-70B-Instruct &  0.5344 & 0.5617 & 0.5367 & 0.5473 & \cellcolor{blue!25}0.4821 & 0.5398  \\
 & gemma-2-9b-it &  0.0902 & 0.0955 & 0.0758 & 0.0856 & 0.0879 & 0.0887  \\
 & gemma-3-27b-it &  0.9332 & \cellcolor{blue!25}0.5686 & \cellcolor{blue!25}0.9105 & \cellcolor{blue!25}0.9037 & \cellcolor{blue!25}0.8915 & \cellcolor{blue!25}0.9173  \\
 & Mixtral-8x7B-Instruct-v0.1 & 0.6285 & 0.6141 & 0.6087 & 0.6201 & 0.5981 & 0.6247  \\
 & Qwen-2.5-72B &  0.5178 & \cellcolor{blue!25}0.4867 & 0.539 & 0.5367 & \cellcolor{blue!25}0.5648 & \cellcolor{blue!25}0.4359 \\

\midrule

\multirow{6}{*}{One-shot}
 & Llama-3.1-8B-Instruct &  0.7513 & \cellcolor{blue!25}0.7126 & \cellcolor{blue!25}0.7126 & \cellcolor{blue!25}0.7088 & \cellcolor{blue!25}0.6974 & \cellcolor{blue!25}0.7081  \\
 & Llama-3.1-70B-Instruct &  0.8597 & \cellcolor{blue!25}0.8544 & \cellcolor{blue!25}0.8961 & \cellcolor{blue!25}0.8529 & \cellcolor{blue!25}0.846 & \cellcolor{blue!25}0.8385  \\
 & gemma-2-9b-it &  0.752 & 0.7528 & 0.7414 & \cellcolor{blue!25}0.7301 & \cellcolor{blue!25}0.7202 & 0.749  \\
 & gemma-3-27b-it &  0.9408 & 0.8574 & 0.915 & 0.9075 & 0.8908 & 0.9175  \\
 & Mixtral-8x7B-Instruct-v0.1 &  0.6406 & \cellcolor{blue!25}0.6178 & \cellcolor{blue!25}0.6187 & \cellcolor{blue!25}0.6148 & \cellcolor{blue!25}0.6103 & \cellcolor{blue!25}0.6269  \\
 & Qwen-2.5-72B & 0.9378 & \cellcolor{blue!25}0.9112 & \cellcolor{blue!25}0.9173 & \cellcolor{blue!25}0.9075 & \cellcolor{blue!25}0.8992 & \cellcolor{blue!25}0.9135 \\

\midrule

\multirow{6}{*}{Zero-Shot CoT}
 & Llama-3.1-8B-Instruct & 0.7467 & 0.7247 & 0.727 &  \cellcolor{blue!25}0.6489 & \cellcolor{blue!25}0.658 & \cellcolor{blue!25}0.6846 \\
 & Llama-3.1-70B-Instruct &  0.8961 & 0.8862 & 0.8809 & \cellcolor{blue!25}0.8597 & \cellcolor{blue!25}0.8605 & \cellcolor{blue!25}0.8741  \\
 & gemma-2-9b-it &  0.7285 & \cellcolor{blue!25}0.7634 & 0.7255 & \cellcolor{blue!25}0.6861 & \cellcolor{blue!25}0.6974 & 0.705 \\
 & gemma-3-27b-it &  0.8612 & \cellcolor{blue!25}0.8218 & \cellcolor{blue!25}0.9067 & \cellcolor{blue!25}0.9052 & \cellcolor{blue!25}0.8893 & \cellcolor{blue!25}0.915  \\
 & Mixtral-8x7B-Instruct-v0.1 &  0.6785 & 0.6694 & 0.6815 & 0.6527 & 0.6573 & 0.6747  \\
 & Qwen-2.5-72B &  0.9393 & \cellcolor{blue!25}0.9075 & \cellcolor{blue!25}0.9158 & \cellcolor{blue!25}0.9105 & \cellcolor{blue!25}0.8991 & \cellcolor{blue!25}0.9203  \\

\midrule

\multirow{6}{*}{One-shot CoT}
 & Llama-3.1-8B-Instruct & 0.7703 & \cellcolor{blue!25}0.7384 & \cellcolor{blue!25}0.332 & \cellcolor{blue!25}0.7263 & \cellcolor{blue!25}0.7414 & \cellcolor{blue!25}0.7361\\
 & Llama-3.1-70B-Instruct &  0.887 & \cellcolor{blue!25}0.8551 & \cellcolor{blue!25}0.8521 & \cellcolor{blue!25}0.8491 & \cellcolor{blue!25}0.8225 & \cellcolor{blue!25}0.8551  \\
 & gemma-2-9b-it & 0.8278 & 0.8082 & 0.818 & 0.8188 & \cellcolor{blue!25}0.796 & 0.8233  \\
 & gemma-3-27b-it & 0.8499 & 0.8165 & \cellcolor{blue!25}0.9128 & \cellcolor{blue!25}0.9044 & \cellcolor{blue!25}0.8893 & \cellcolor{blue!25}0.9112 \\
 & Mixtral-8x7B-Instruct-v0.1 &  0.6876 & 0.6846 & \cellcolor{blue!25}0.655 & 0.6648 & \cellcolor{blue!25}0.6588 & 0.6618 \\
 & Qwen-2.5-72B & 0.9408 & 0.9112 & 0.9188 & 0.9105 & 0.9029 & 0.9234 \\

\midrule

\multirow{6}{*}{COD}
 & Llama-3.1-8B-Instruct & 0.7058 & 0.6921 & 0.7073 & 0.7096 & 0.6891 & 0.7081  \\
 & Llama-3.1-70B-Instruct &  0.8627 & \cellcolor{blue!25}0.8483 & \cellcolor{blue!25}0.846 & \cellcolor{blue!25}0.8385 & \cellcolor{blue!25}0.8362 & \cellcolor{blue!25}0.8369 \\
 & gemma-2-9b-it & 0.1698 & 0.1978 & 0.1675 & 0.1751 & 0.1523 & 0.169  \\
 & gemma-3-27b-it &  0.8923 & \cellcolor{blue!25}0.8711 & \cellcolor{blue!25}0.8749 & \cellcolor{blue!25}0.8695 & \cellcolor{blue!25}0.8559 & 0.8832  \\
 & Mixtral-8x7B-Instruct-v0.1 &  0.4912 & \cellcolor{blue!25}0.5109 & \cellcolor{blue!25}0.5087 & \cellcolor{blue!25}0.5034 & \cellcolor{blue!25}0.4692 & 0.4761 \\
 & Qwen-2.5-72B &  0.8733 & 0.8551 & 0.8536 & 0.8544 & 0.8529 & 0.8658  \\

\bottomrule
\end{tabular}
\end{adjustbox}
\caption{LLM Performance on GSM8K Dataset Across Cultures and Prompting Techniques. Blue cell indicates a statistically significant ($p < 0.05$) difference in accuracy compared to the corresponding US baseline.}
\label{tab:results-num}
\end{table*}

\section{Resources}
All experiments in this study were conducted using NVIDIA A100 GPUs, with each model evaluated across multiple cultural variants and prompting strategies. The culturally adapted GSM8k dataset comprises 7,914 verified math problems spanning six cultural contexts. Manual verification was performed by a team of trained annotators to ensure semantic and numerical fidelity across adaptations (annotator details provided in Appendix \ref{app:annotator}). In total, we evaluated six open-source large language models using a combination of zero-shot, one-shot, and chain-of-thought prompting techniques. Our codebase, evaluation scripts, and datasets will be made publicly available to support reproducibility and further research.
\end{document}